\pdfoutput=1

\documentclass[11pt]{article}

\usepackage[final]{acl}

\usepackage{times}
\usepackage{latexsym}

\usepackage[T1]{fontenc}

\usepackage[utf8]{inputenc}

\usepackage{microtype}

\usepackage{inconsolata}

\usepackage{graphicx}
\usepackage{paralist}
\usepackage{array}
\usepackage{booktabs}
\usepackage{colortbl}
\usepackage{hyperref}
\usepackage{authblk}
\usepackage{tablefootnote}

\usepackage[normalem]{ulem}

\def\ODdel#1{\bgroup\markoverwith{\textcolor{purple!60}{\rule[0.4ex]{2pt}{3pt}}}\ULon{#1}}

\def\SMdel#1{\bgroup\markoverwith{\textcolor{green!60}{\rule[0.4ex]{2pt}{3pt}}}\ULon{#1}}

\def\OPdel#1{\bgroup\markoverwith{\textcolor{pink!60}{\rule[0.4ex]{2pt}{3pt}}}\ULon{#1}}

\def\PSdel#1{\bgroup\markoverwith{\textcolor{magenta}{\rule[0.4ex]{2pt}{3pt}}}\ULon{#1}}

\def\DMHdel#1{\bgroup\markoverwith{\textcolor{blue!60}{\rule[0.4ex]{2pt}{3pt}}}\ULon{#1}}

\definecolor{overlap}{HTML}{c7522a}
\definecolor{semsim}{HTML}{e5c185}
\definecolor{textprop}{HTML}{fbf2c4}
\definecolor{textclf}{HTML}{b8cdab}
\definecolor{match}{HTML}{74a892}
\definecolor{perplexity}{HTML}{008585}
\definecolor{factuality}{HTML}{4c9eb3}
\definecolor{distance}{HTML}{779af5}
\definecolor{combination}{HTML}{a59cff}
\definecolor{infspeed}{HTML}{dbcdf0}

\title{Automatic Metrics in Natural Language Generation: \\ A Survey of Current Evaluation Practices}

\usepackage[misc]{ifsym}
\author[1 {\normalfont\Letter}]{\bf Patrícia Schmidtová}
\author[2]{\textbf{Saad Mahamood}}
\author[1]{\textbf{Simone Balloccu}}
\author[1]{\\ \textbf{Ondřej Dušek}} 
\author[3]{\textbf{Albert Gatt}}
\author[4]{\textbf{Dimitra Gkatzia}}
\author[4]{\\ \textbf{David M. Howcroft}}
\author[1]{\textbf{Ondřej Plátek}}
\author[5]{\textbf{Adarsa Sivaprasad}}

\affil[1]{Charles University, Faculty of Mathematics and Physics, Prague, Czechia}
\affil[2]{trivago N.V., Düsseldorf, Germany}
\affil[3]{Utrecht University, Utrecht, Netherlands}
\affil[4]{Edinburgh Napier University, Edinburgh, Scotland, United Kingdom}
\affil[5]{University of Aberdeen, Aberdeen, Scotland, Untied Kingdom}
\affil[ ]{{\normalfont\Letter} Corresponding author: \href{mailto:schmidtova@ufal.mff.cuni.cz}{schmidtova@ufal.mff.cuni.cz}}

\begin{document}
\maketitle
\begin{abstract}
Automatic metrics are extensively used to evaluate natural language processing systems. 
However, there has been increasing focus on how they are used and reported by practitioners within the field. 
In this paper, we have conducted a survey on the use of automatic metrics, focusing particularly on natural language generation (NLG) tasks. We inspect which metrics are used as well as why they are chosen and how their use is reported. Our findings from this survey reveal significant shortcomings, including inappropriate metric usage, lack of implementation details and missing correlations with human judgements. We conclude with recommendations that we believe authors should follow to enable more rigour within the field. 
\end{abstract}

\section{Introduction}
\label{sec:introduction}

Evaluation practices in the field of Natural Language Processing (NLP) are increasingly coming under a microscope by researchers. 
There is now a significant body of contributions presenting experimental research, meta-analyses and/or best practice guidelines, on issues ranging from statistical significance testing~\cite{Dror2017}, to human evaluation methods~\cite{XXXhowcroft-etal-2020-twenty,XXXvan-der-lee-2021-best-practices,hamalainen_human_2021,XXXshimorina-belz-2022-human}, error analysis~\cite{XXXvan-miltenburg-etal-2021-underreporting,van-miltenburg-etal-2023-barriers} and replicability of evaluations~\cite{XXXbelz-etal-2021-systematic,belz-etal-2023-missing}.

Automatic metrics and their usage for evaluation have also been under extensive examination by researchers. 
Similarity-based metrics are sometimes taken as proxies for human quality ratings, whereas findings suggest the two should not be conflated. This has lead to concerns about metric validity \cite{belz-gatt-2008-intrinsic}. For example, the validity of metrics such as BLEU \cite{papineni-etal-2002-bleu} and ROUGE \cite{lin-2004-rouge} has been put into question regarding their poor correlation with human judgements \cite{reiter-belz-validity-2009, novikova-etal-2017-need, reiter-2018-structured}. 
In addition, automatic metrics do not capture factuality or faithfulness issues in text \cite{gehrmann-2023-repairing}, such as incorrect names and numbers \cite{thomson-reiter-2020-gold}.
Interpreting the meaning of scores generated by automatic metrics can also be challenging. 
For example, what researchers often report as a ``BLEU score'' actually consists of several metrics, depending on multiple parameters and varying across different implementations, which are not compatible with each other \cite{post-2018-call}.
There are also questions on whether it is possible to encapsulate the performance of a given system with a single number or whether the use of a single metric to demonstrate improvements over prior systems provides sufficient dimensionality in reporting the performance characteristics of a given system \cite{gehrmann-2023-repairing}.   

Given the well-documented shortcomings of automatic metrics, our goal in this paper is to survey the current state of play in  metric-based evaluations of natural language generation (NLG). As with the above-mentioned studies focusing on other facets of evaluation, we aim to both understand how metrics are currently used in NLG, and to identify gaps and possible ways forward in an effort to improve the scientific quality of NLG research.

Specifically, we conduct an analysis of published work in the field, annotating which metrics are used, for what purposes, and how their usage is reported. In Section~\ref{sec:background}, we describe past survey efforts within the field of NLG to frame our contribution.  In Section~\ref{sec:method}, we describe our paper selection procedure, the annotation procedure, the challenges we encountered, and the process and results of computing inter-annotator agreement between the annotators. The analysis and results from the annotation process are presented in Section~\ref{sec:results}, and we offer our insights into these results in Section~\ref{sec:discussion}. Finally, we wrap up with recommendations (Section~\ref{sec:recommendations}) and concluding thoughts (Section~\ref{sec:conclusions}) from the observations based on our results. 

\section{Evaluation Surveys in NLG}
\label{sec:background}

There have been several surveys  inspecting the different aspects of evaluation practices within NLG over the last several years. Some surveys focused on quantifying the types of evaluations, the proportion of intrinsic and extrinsic evaluations over a defined period of time either for the field as a whole \cite{gkatzia-mahamood-2015-snapshot}, or for a specific domain such as question generation \cite{amidei-etal-2018-evaluation}.  In addition, there has been an effort to understand the different types of metrics and evaluation approaches employed and to categorise the challenges faced by researchers \cite{celikyilmaz-evaluation-2020}.

In addition to survey work covering shortcomings of automatic metrics \cite{gehrmann-2023-repairing}, a significant amount of work has focused on human evaluation practices within NLG. Past work has revealed a large variation in practices among researchers \cite{van-der-lee-etal-2019-best}. This was followed up by an extensive survey which has shown that in addition to the large variety of practices, there are fundamental gaps in reported details by authors \cite{howcroft-etal-2020-twenty}. These issues have led to proposals for best practices for carrying out and reporting human evaluations in NLG \cite{van-der-lee-2021-best-practices, shimorina-belz-2022-human}. However, the concern about human evaluation practices has also led researchers to consider whether human evaluations in NLG -- and in NLP as a whole -- are both reproducible and repeatable \cite{belz-etal-2023-non} given the inconsistencies and gaps in reporting practices.  

One area where reporting practices have received attention is the way in which errors made by NLG systems are documented. 
\Citet{van-miltenburg-etal-2021-underreporting} found that there is severe under-reporting of the different kinds of errors a given NLG system can make, which leaves the broader community ``in the dark'' due to this missing information. 
Beyond evaluations and reporting practices, there have been attempts to better understand the motivations of researchers and their reporting practices by directly surveying them. 
\citet{zhou-etal-2022-deconstructing} found that there is pressure towards a ``kitchen sink'' approach for evaluation. 
Even though researchers recognise the limitations of existing metrics, lack of clarity about their evaluation goals and quality criteria can lead to over-reporting potentially uninformative metrics \cite{zhou-etal-2022-deconstructing}. 
Other work explored the barriers that researchers face to conducting error analyses \cite{van-miltenburg-etal-2023-barriers}: 
while respondents were generally positive about error analyses, there are multiple barriers such as page limits, lack of tools or resources, and a lack of time and/or money.       

\section{Survey Method}
\label{sec:method}

Although past surveys looked at the deficiencies of automatic metrics, none of them go beyond quantifying and aggregating their usage. This is necessary, considering that the use of automatic metrics increased by almost 25\% in the 2016-2019 period, with some surveys reporting that almost half of the papers surveyed only use automatic metrics \cite{van-der-lee-2021-best-practices}. To obtain a comprehensive and up-to-date view of current practices in automatic evaluation for NLG, we have focused on recently published articles in prominent, peer-reviewed venues.

\paragraph{Paper selection} Our analysis is based on a snapshot of a total of 110 papers presented in 2023 in two relevant venues: the {\em International Conference on Natural Language Generation} (INLG) and the {\em Annual Meeting of the Association for Computational Linguistics} (ACL). All papers ($n=36$, of which $26$ are long papers) at the main conference track of INLG 2023 were included. For ACL, we used all the papers presented under the {\em Generation} track ($n=74$, $63$ are long papers). In addition to regular ACL papers, this included three papers originally accepted for publication in the {\em Transactions of the ACL} (TACL) journal and one NLG paper from the journal {\em Computational Linguistics}. %

\begin{table*}[!h]
    \centering
    \small
    \begin{tabular}{p{2.55cm}p{9.6cm}cc}
    \toprule
        {\bf Feature} &  {\bf Description} & {\bf IAA (J)} & {\bf IAA (M)} \\
        \midrule
        Name & Which evaluation method was used? (Appendix \ref{app:metrics}).  & 0.59 & 0.34 \\
        Newly introduced? & Was the metric newly introduced in this paper? & 0.76 & 0.76 \\
        Task & Which task(s) this metric was used to evaluate (Appendix \ref{app:tasks})? & 0.41 & 0.35 \\
        Human Correlations & Were automatic metric results directly related to human evaluation results? Was this correlation quantitative or qualitative? & 0.47 & 0.47 \\
        Implementation & Were specific metric implementation details (e.g. links to the specific metric implementation, paper reference, etc.) provided or not?  & 0.44 & \phantom{$^1$}0.45\footnotemark  \\
        Appendix & Was the metric only reported in the Appendix, rather than the main section of the paper? & 0.61 & 0.60\\
         Rationale & Did the authors explain the rationale for the metric? & N/A & N/A\\
         \bottomrule
    \end{tabular}
    \caption{Features annotated for each paper. IAA (J/M): inter-annotator agreement between 6 authors for 4 papers on each criterion, using the Jaccard or MASI distance metrics.\textsuperscript{\ref{fn:iaa}} Note that `Rationale' is not included in the agreement computation since it was recorded in a free-text form to allow for more flexibility.}
    \label{tab:annotation}
\end{table*}

\paragraph{Annotation procedure} Papers were randomly distributed among all the authors in a set of annotation batches, and independently annotated for the features summarised in Table \ref{tab:annotation}. 
As the table indicates, the main purpose of the annotation was to identify which {\em automatic} evaluation metrics or {\em human} evaluation methods (if any) are reported in the paper and for which tasks. 
A full list of evaluation methods identified is provided in Appendix \ref{app:metrics}. 
We annotated the task type using definitions created by \citet{howcroft-etal-2020-twenty}; annotators could also include other tasks not in this list if necessary 
(see Appendix \ref{app:tasks} for details). 
Note that it is possible for papers to report different metrics for different evaluation experiments, depending on the (sub)task.
Crucially, we also consider whether a metric is newly introduced in a paper or was previously published. In either case, we are interested in whether authors describe the rationale for their use of a metric. In case a paper included a human evaluation, we also annotate whether metric-based evaluations were quantitatively correlated with the outcomes of the human evaluation, or whether there was any qualitative discussion of the relationship between the two.

\footnotetext[1]{Note that these correlation values relate to the initial annotation guidelines and the \emph{link to metric} property, which directly compared implementation URLs and was not clear on the procedure if a paper used multiple implementations. The feature was then changed into the categorical \emph{implementation details} with three options: no implementation details provided, implementation details provided, and multiple implementations used.}

\paragraph{Iterative refinement and inter-annotator agreement} Annotation proceeded in multiple rounds. During an initial round, we independently annotated a subset of papers and discussed the outcomes to fine-tune the annotation scheme. Subsequently, a random sample of 4 papers (2 from INLG; 2 from ACL) was selected and independently annotated by 6 of the authors. %
Inter-annotator agreement (IAA) for the features outlined in  Table~\ref{tab:annotation} was computed using both the Jaccard and MASI \cite{Passonneau2006} distance metrics.\footnote{We estimate agreement using the {\tt AnnotationTask} class and {\tt jaccard\_distance} and {\tt masi\_distance} functions in the NLTK {\tt metrics} library \cite{BirdKleinLoper09}.\label{fn:iaa}} 
Following discussions, we addressed the disagreements by replacing the originally annotated \emph{link to metric} with \emph{implementation details} and reporting \emph{task} using a selection from a drop-down list following \citet{howcroft-etal-2020-twenty}'s definitions. %

\section{Analysis and Results}
\label{sec:results}

We present the results of our annotation of 110 papers in this section. 
Out of the 110 papers annotated, a total of 102 papers included an evaluation of a generation system. The excluded 8 papers did not propose any systems to be evaluated. For example, they either presented a corpus or methods to detect the decoding algorithm of a closed-source model. After the removal of these papers, a total of 69 ACL papers and 33 INLG papers were analysed. 

A total of 59 papers (56.73\%) of papers use human evaluations; in contrast, 98 papers (94.23\%) used automatic metrics, a result similar to what was found by \citet{van-der-lee-2021-best-practices}, who reported 95\% of papers using automatic metrics in both ACL tracks and INLG. There were only 53 papers (50.96\%) containing both automatic and human evaluation. 

Another aspect explored was whether authors provide any implementation details, such as link to the specific implementation used for the evaluation. We found that for 66.2\% of INLG and 57.3\% of ACL papers, these details were not mentioned either in the main body of the paper or within the appendices. 
Given the high percentage of papers not giving specific implementation details, this can make it difficult to conduct reproduction studies under the same conditions, especially, considering how challenging it is to reproduce the original scores of NLP evaluations \cite{belz-etal-2021-systematic}.

In the subsequent sections, we will explore in more detail how specific metrics are used (Section~\ref{subsec:metric-analysis}), what the relationship is between automatic and human evaluations (Section~\ref{subsec:automatic_and_human_eval}), how these relate to different NLG subtasks  (Section~\ref{subsec:task_repersentation}), and whether the papers provide their code (Section~\ref{subsec:paper_resource_findings}), an important consideration given the concern about evaluation reproducibility.

\subsection{Metric-Level Analysis}
\label{subsec:metric-analysis}

\begin{table}[t]
\centering
\scriptsize
\begin{tabular}{lrrrl}
\toprule
\textbf{Metric Family Name}  & \textbf{INLG} & \textbf{ACL} & \textbf{Total} &  \\ 
\midrule
\rowcolor{overlap!60} BLEU                  &26                  &69                  &95             &  \\ 
\rowcolor{overlap!60} ROUGE                  &27                  &65                  &92             &  \\ 
\rowcolor{textprop!60} N-gram diversity                  &6                  &49                  &55             &  \\ 
\rowcolor{textclf!60} Style Classifier                  &5                  &37                  &42             &  \\ 
\rowcolor{semsim!60} BERTScore                  &8                  &32                  &40             &  \\ 
\rowcolor{perplexity!60} Perplexity                  &3                  &29                  &32             &  \\
\rowcolor{overlap!60} METEOR                  &6                  &21                  &27             &  \\ 
\rowcolor{semsim!60} Semantic Similarity                  &9                  &12                  &21             &  \\ 
\rowcolor{overlap!60} Overlap                  &6                  &21                  &27             &  \\
\rowcolor{factuality!60} Factuality                  &5                  &13                  &18             &  \\
\rowcolor{match!60} Accuracy                  &8                  &8                  &16             &  \\ 
\rowcolor{textclf!60} Quality Estimation                  &7                  &7                  &14             &  \\  
\rowcolor{magenta!60} Combination                  &0                  &14                  &14             &  \\
\rowcolor{semsim!60} BARTScore                  &2                  &10                  &12             &  \\ 
\rowcolor{factuality!60} NLI                  &44                  &8                  &12             &  \\  
\rowcolor{match!60} F1                  &4                  &7                  &11             &  \\ 
\rowcolor{textclf!60} BLEURT                  &5                  &5                  &10             &  \\ 
\rowcolor{overlap!60} CIDEr                  &2                  &6                  &8             &  \\ 
\rowcolor{textprop!60} N-gram repetition                  &2                  &6                  &8             &  \\ 
\rowcolor{overlap!60} SARI                  &2                  &6                  &8             &  \\ 
\rowcolor{textprop!60} Sequence Length                  &3                  &5                  &8             &  \\ 
\rowcolor{perplexity!60} MAUVE                  &0                  &8                  &8             &  \\ 
\rowcolor{textclf!60} Unieval                  &0                  &8                  &8             &  \\ 
\rowcolor{distance!60} Distribution Comparison                  &0                  &7                  &7             &  \\ 
\rowcolor{overlap!60} NIST                  &0                  &7                  &7            &  \\
\rowcolor{semsim!60} MoverScore                  &1                  &5                  &6             &  \\ 
\rowcolor{overlap!60} PARENT                  &1                  &5                  &6             &  \\ 
\rowcolor{match!60} Recall                  &2                  &44                  &6             &  \\ 
\rowcolor{distance!60} Edit Distance                  &1                  &5                  &6             &  \\ 
\rowcolor{textprop!60} Flesch Readability                  &1                  &3                  &4             &  \\ 
\rowcolor{infspeed!60} Inference Speed                  &0                  &4                  &4             &  \\
\rowcolor{match!60} Precision                  &1                  &2                  &3             &  \\ 
\rowcolor{distance!60} loss/error                  &0                  &3                  &3             &  \\ 
\rowcolor{overlap!60} chrF++                  &1                  &1                  &2             &  \\ 
\bottomrule
\end{tabular}
\caption{Total automatic metric usage counts of each of the metric families for both INLG and ACL conferences.}
\label{tab:metric_family_counts}
\normalsize
\end{table}

\begin{table}[t]
\centering
\scriptsize
\begin{tabular}{lrrrl}
\toprule
\textbf{Metric Task Name} & \textbf{INLG} & \textbf{ACL} & \textbf{Total} &  \\ 
\midrule
\rowcolor{overlap!60}Overlap&71&201&272 &  \\
\rowcolor{semsim!60}Semantic Similarity&20&59&79 &  \\
\rowcolor{match!60}Match&15&61&76 &  \\
\rowcolor{textprop!60}Text Properties&12&63&75 &  \\
\rowcolor{textclf!60}Text Classifier&17&57&74 &  \\
\rowcolor{factuality!60}Factuality&49&21&70 &  \\
\rowcolor{perplexity!60}Perplexity&3&37&40 &  \\
\rowcolor{distance!60}Distance-based&1&15&16 &  \\
\rowcolor{combination!60}Combination&0&14&14 &  \\
\rowcolor{infspeed!60}Inference Speed&0&4&4 &  \\
\bottomrule
\end{tabular}
\caption{Total usage counts of each of the high-level metric categories for both INLG and ACL conferences.}
\label{tab:metric_category_counts}
\normalsize
\end{table}

\begin{figure}
    \centering
    \includegraphics[width=\linewidth]{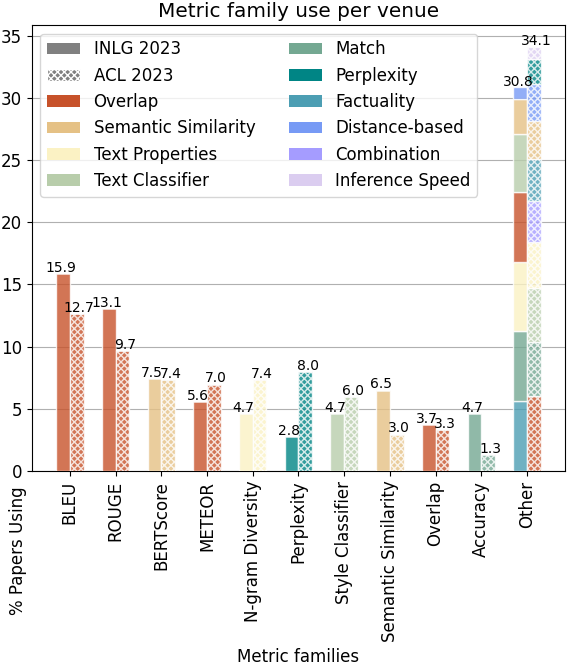}
    \caption{Usage percentages of top 10 metric families in INLG and ACL, with metric category color-coded.}
    \label{fig:metric_family_usage_across_venues}
\end{figure}

We identified 634 counts of automatic metric uses within these papers, with 283 different automatic metric names used by practitioners. %
To enable further analysis of these metrics %
and to derive useful insights into researcher practices, we manually grouped the metrics into 38 \emph{metric families} that group together similar metrics. 
In particular, we aimed at the most informative grouping possible: We merged similar metrics which are individually relatively rare, while keeping frequently used metrics within their own family (e.g., BLEU).
We further joined the metric families into 10 broad \emph{metric categories} to enable a more high-level overview.
Table~\ref{tab:metric_category_counts} lists all metric categories with their usage counts across the surveyed papers. 
Table~\ref{tab:metric_family_counts} shows the number of metric occurrences in papers across metric families, with colour codes corresponding to metric categories in Table~\ref{tab:metric_category_counts}. %
The overall usage of the most frequent metric families and the corresponding categories is further depicted in Figure~\ref{fig:metric_family_usage_across_venues}.
The full list of all identified metrics and their grouping can be found in Appendix~\ref{app:metrics}.

\begin{figure}[t]
    \centering
    \includegraphics[width=\linewidth]{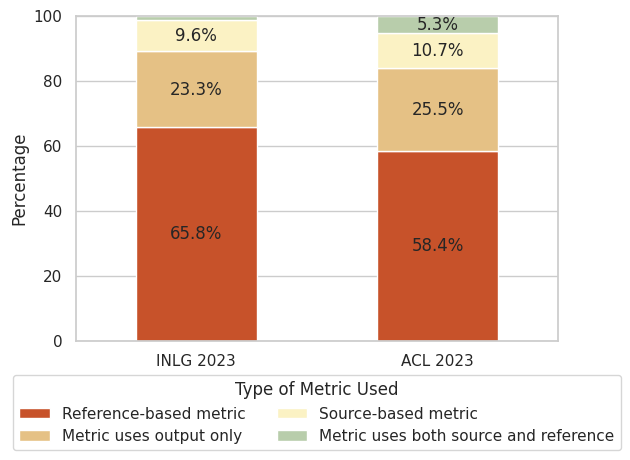}
    \caption{The percentage of automatic metric types used in both INLG and ACL conferences.}
    \label{fig:src_ref}
\end{figure}

\paragraph{Frame of comparison:}
We further divide metrics into \textit{reference-based} (use a human reference or pairwise output from another system), \textit{source-based} (mostly checking for output fidelity/alignment with the input), \textit{output only} (evaluating inherent text properties such as diversity), or \textit{source and reference based}. We find that the dominant form is reference-based metrics: As show in Figure~\ref{fig:src_ref}, this holds true in both INLG and ACL papers, with this metric type used more extensively in INLG compared to ACL. This suggests that researchers are primarily looking to evaluate the outputs of systems against reference corpora to get an estimation of performance. Some metrics, such as SelfBLEU, can be used in multiple different ways, which may inflate the usage estimates for reference-based metrics.

\paragraph{BLEU and ROUGE:}
Across both INLG and ACL papers, BLEU and ROUGE are the predominant metrics used for NLG automatic evaluations, as seen in Table~\ref{tab:metric_family_counts}. 
This is in line with previous qualitative observations \cite{van-der-lee-2021-best-practices, gehrmann-2023-repairing}. 
Interestingly, as shown in Figure~\ref{fig:metric_family_usage_across_venues}, the usage of BLEU and ROUGE is proportionately higher in INLG compared to the ACL Generation track. 
BLEU is the most popular metric in both INLG and ACL, despite the multiple concerns raised by researchers on its validity as an NLG metric \cite{reiter-2018-structured}. 
Moreover, for 63.6\% of papers using BLEU and 62.6\% of those using ROUGE no implementation details were provided, despite the compatibility issues this can cause \cite{post-2018-call,grusky-2023-rogue}.

\paragraph{Trainable metrics} (mostly from the Semantic Similarity, Text Classifier, and Factuality categories)
only make up a minority, with 28.4\% in INLG and 35.5\% in ACL, respectively. This suggests that even though learning-based metrics such as BERTScore \cite{zhang-2019-bertscore}, BLEURT \cite{sellam-2020-bleurt}, etc.\ are gaining traction, they are still not as popular as more basic approaches. %

\begin{figure}[t]
    \centering
    \includegraphics[width=\linewidth]{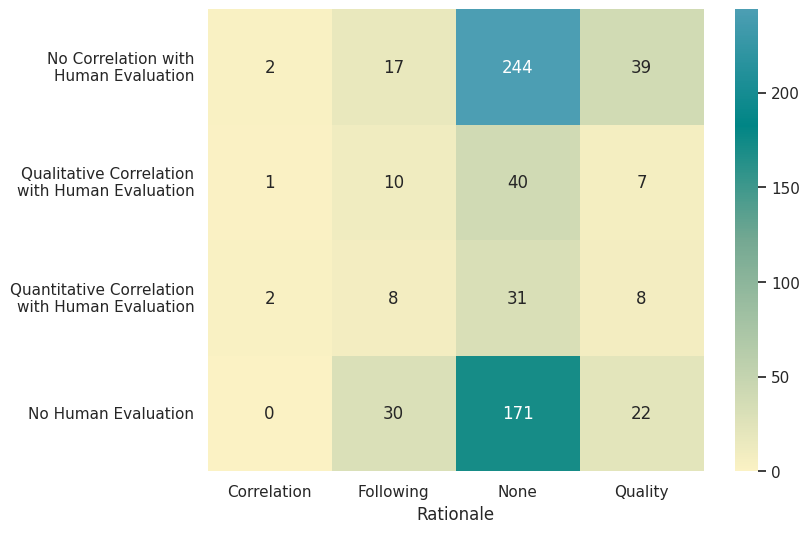}
    \caption{Co-occurrence of types of rationales with the authors correlating the metric results to human judgment. }
    \label{fig:humeval_vs_rationale}
\end{figure}

\paragraph{Metric Rationales:}
The vast majority of annotated metrics (486, 76.9\%) did not include a rationale for the use of a metric A total of 65 mentions of metrics in papers (10.3\%) stated that they were following previous work. 
Authors rationalized five of the metrics by stating that they correlate with human judgements, generally shown by previous work. Finally, for 76 metrics (12.0\%), a rationale other than following previous work or correlating with human judgement was stated in the papers, e.g. that the given metric was included to measure fluency or diversity.%

We also looked at the relationship between the type of rationale given for a metric and whether a correlation with human evaluation was discussed (Figure~\ref{fig:humeval_vs_rationale}). It is very clear that for a vast majority of metrics no rationale is provided, irrespective of whether a human evaluation has been conducted or not.

\paragraph{New Metrics:}
We found that 76 new metrics were introduced, with eight of them named and proposed for future use:
\begin{compactitem}
    \item AlignScore \cite{zha-etal-2023-alignscore}
    \item NegBleurt \cite{anschutz-etal-2023-correct}
    \item NegMPNet \cite{anschutz-etal-2023-correct}
    \item HAUSER \cite{he-etal-2023-hauser}
    \item WeCheck \cite{wu-etal-2023-wecheck}
    \item NEHR \cite{akani-etal-2023-reducing}
    \item LENS \cite{maddela-etal-2023-lens}
    \item DecompEval \cite{ke-etal-2023-decompeval}
\end{compactitem}
All of these metrics are based on trainable components and mostly focus on factual correctness, going against the majority currently in use, but reflecting an emerging trend.
It would be interesting to observe in the future whether these new metrics are adopted by the research community or not.  

\paragraph{Appendix: } We observed that, for a given paper, some metrics are only reported in the papers' appendices. %
This was the case for nine metrics (4.8\%) at INLG and 22 metrics (3.8\%) at ACL.

\subsection{Automatic vs.~Human Evaluations}
\label{subsec:automatic_and_human_eval}

\begin{figure}
    \centering
    \includegraphics[width=\linewidth]{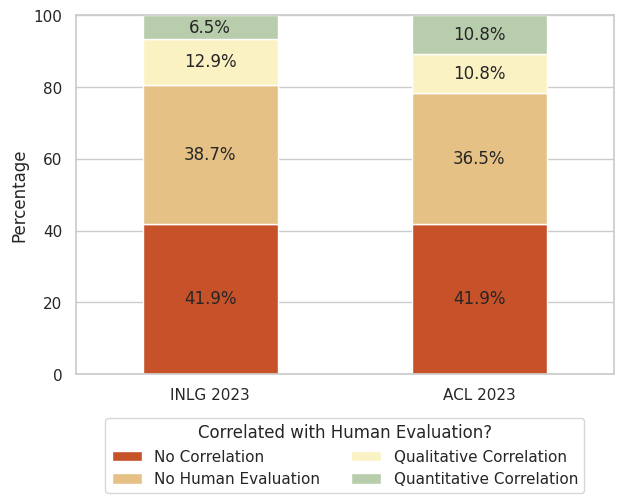}
    \caption{The percentage of papers that state a form of correlation between their automatic and human evaluation results.}
    \label{fig:corr_human}
\end{figure}

We conducted an additional analysis to better understand whether researchers treat their automatic and human evaluations as separate entities, or seek a more unified interpretation of results from the two, by looking for correlations between them. We annotated papers with four approaches to their human evaluations:

\begin{compactitem}
    \item \textit{Quantitative Correlation} - Cases where the authors check if  their automatic metric result(s) quantitatively correlate with evaluation results from their own or previous work. 
    \item \textit{Qualitative Correlation} - When authors only draw qualitative conclusions on the relation between their automatic and human evaluation results, without statistical analysis to back this claim.
    \item \textit{No Correlation} - No stated correlation either quantitatively or qualitatively can be found in the paper.
    \item \textit{No Human Evaluation} - No evaluation involving human participants was performed by the researchers. 
\end{compactitem}

Interestingly, papers from the ACL generation track and INLG are very similar in terms of correlating with human evaluations, as shown in Figure~\ref{fig:corr_human}. Papers predominantly either did not perform a human evaluation or if they did, they did not check for a correlation between their automatic and human evaluation results. 
Authors who provided either a qualitative or quantitative analysis between their automatic and human evaluation results are very much in the minority. 

A possible reason for the low level of reported correlations between automatic and human evaluations could be the known issues between lexical overlap evaluation metrics and specific NLG sub-tasks, such as referring expression generation \cite{belz-gatt-2008-intrinsic}. 
An alternative possibility is that while automatic metrics may give an approximate estimate of language quality, they do not measure content quality \cite{reiter-belz-validity-2009} and therefore researchers are looking to measure different aspects with their automatic and human evaluations.

\subsection{Task Representation}
\label{subsec:task_repersentation}

\begin{table}[t]
\centering
\footnotesize
\addtolength{\tabcolsep}{-0.4em}
\begin{tabular}{lrrr}
\toprule
\textbf{Task Name} & \textbf{INLG} & \textbf{ACL} & \textbf{Total}   \\ 
\midrule
Summarisation (text-to-text)& 6 & 17 &  23  \\
Feature-Controlled Generation & 5 & 13 & 18   \\
Dialogue Turn Generation & 3 & 10 & 13  \\
Data-to-text Generation & 5 & 8 & 12   \\
Machine Translation & 0 & 10 &  10   \\
Question Generation & 1 & 9 & 10 \\
Paraphrasing/Lossless Simplification & 1 & 9 & 10 \\
Question Answering & 0 & 8 & 8 \\
End-to-End Text Generation & 1 & 7 & 8 \\
Story Generation & 3 & 3 & 6 \\
LM Sampling & 2 & 3 & 5 \\
Referring Expression Generation & 2 & 0 & 2 \\
Content Selection/Determination & 1 & 2 & 3 \\
Surface Realisation (SLR to Text) & 0 & 2 & 2 \\
Song Lyric Generation & 0 & 2 & 2  \\
Compression/Lossy Simplification & 0 & 2 & 2 \\
Commonsense Reasoning & 0 & 2 & 2 \\
Aggregation & 0 & 1 & 1 \\
\bottomrule
\end{tabular}
\caption{List of NLG task types, with counts of relevant papers from the annotated sets. Task definitions are based on those used by \citet{howcroft-etal-2020-twenty}.}
\label{tab:metric_task_counts}
\normalsize
\end{table}

\begin{figure*}
    \centering
    \includegraphics[width=\linewidth]
    {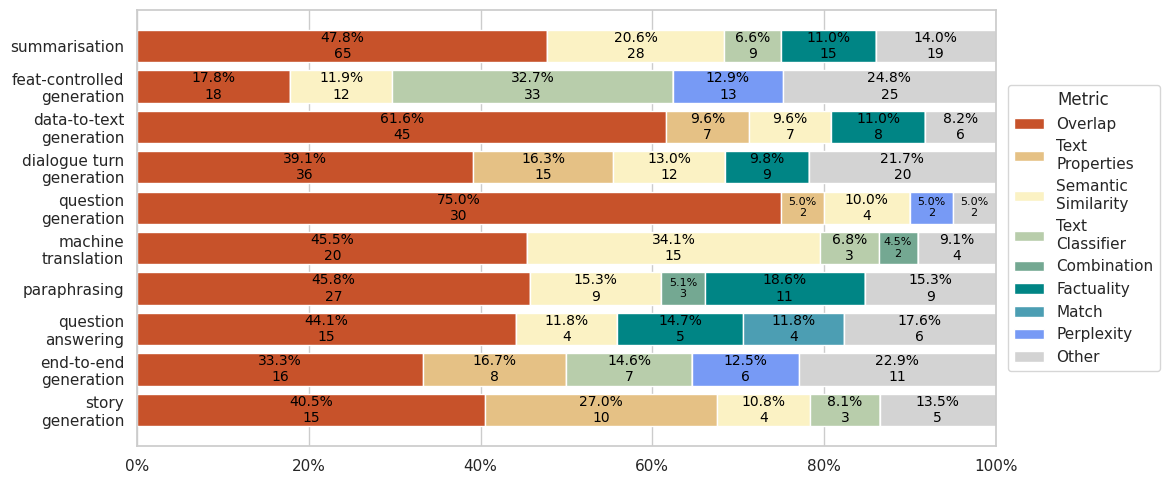}
    \caption{Distributions of different metric families used to evaluate a given task across ACL and INLG (with percentages of metric usages for the given task on top and absolute counts below).}
    \label{fig:metric_task_usage}
\end{figure*}

Table~\ref{tab:metric_task_counts} shows the counts for each of the task types, with the majority of papers focusing on text-to-text summarisation. 
We analysed the relationship between the paper task and metric usage, shown in Figure~\ref{fig:metric_task_usage}. 
Overlap metrics dominate most tasks, especially question generation (75\%) and data-to-text generation (61.6\%). 
Interestingly, feature-controlled generation seems to be the only task that sees some of the lowest usage of 
Overlap metrics (17.8\%); moreover, in comparison to other tasks it is the only one where other metrics are dominant. %

\subsection{Paper Resources Findings}
\label{subsec:paper_resource_findings}
Our last area of analysis was the completeness of paper code resources. Given the importance of complete code and resources for the reproduction and repeatability of experiment results, we manually checked papers to see not only if they provided a link to an implementation, but also if the given link contained any code or data.

\paragraph{Annotation approach:} We classified papers into three groups: \textit{delivered} if the code was present, \textit{no code} if not and the paper did not promise any code, and finally \textit{missing}, which applied to papers that linked to code repositories, but these were either dead, empty or contained only abstracts or titles or promises of a future release. For papers that delivered code, we also annotated the following aspects (see appendix D.1 for more details):

\begin{compactitem}
        \item Granularity of installation instructions: \textit{None, Basic, Detailed}
        \item Clarity of experiments structure in the code, whether experiments described in the papers are ``discoverable'': \textit{None, Some, Many}
        \item Level of documentation detail, such as if hyperparameters are described and how experiments can be executed: \textit{None, Basic, Detailed}
\end{compactitem}

\paragraph{Code availability:} In terms of available code, 75\% of INLG and 70.2\% of ACL papers published their code. 18.2\% INLG papers and 11.9\% ACL papers published no code. This is similar to the results of \citet{mieskes-2019-nlp}, who found no code in 14\% cases and no experimental resources in 11.1\% cases. A larger proportion of ACL papers (17.9\%) promised to deliver code but did not, compared to 6.8\% for INLG.
We examined the papers annotated as \textit{missing} to further understand if there was a difference between authors who come from industry as compared to academia. Papers were classified as being ``industry'' papers if a majority of author affiliations are not from an academic institution. We found that the majority of \textit{missing} papers have either complete or partial industry authorship ($n=13$), compared to purely academia papers ($n=5$). Whilst the numbers detected are too small to draw definite conclusions, we hypothesise that additional business constraints increase the likelihood of not releasing the code even if promised by the authors.   

\paragraph{Examining code releases:} For papers that had published code, we considered the level of detail of the installation instructions provided. For 52.7\% of ACL papers and 50\% of INLG papers, no installation instructions were provided. For the remaining papers, 13.9\% and 10.8\% of INLG and ACL papers respectively provided basic installation instructions. This leaves a minority of 36.1\% and 36.5\% of INLG and ACL papers with detailed installation instructions. 

A similar story holds for how discoverable experiments are within papers that have published code. In only a minority of papers (27.8\% for INLG and 37.8\% for ACL), half or more of their experiments could be directly linked to scripts provided within the code. 

In terms of code documentation, an alarming 44.4\% of INLG and 43.2\% of ACL paper resources provide no instructions whatsoever.  

\begin{figure}
    \centering
    \includegraphics[width=\linewidth]{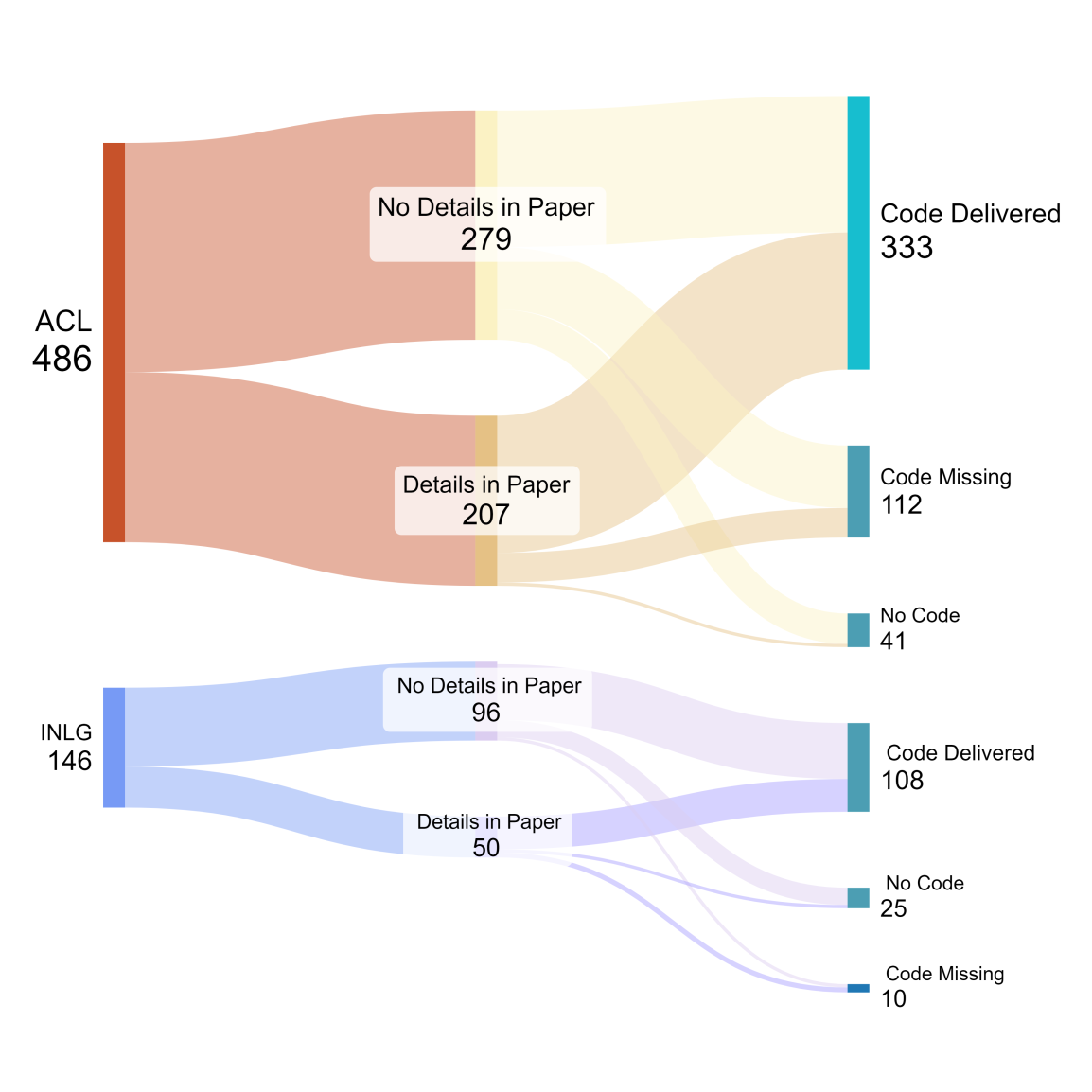}
    \caption{The proportion of metrics across ACL and INLG and the availability of paper resources.}    \label{fig:sankey_metrics_resources}
\end{figure}

\paragraph{Metrics and Paper Resources:} We also explored the relationship between inclusion of metric implementation details in a paper and the availability of paper resources. Figure~\ref{fig:sankey_metrics_resources} shows a visualisation of this analysis. The main point that stands out is that for metrics with no implementation details, there is a larger proportion of papers with missing code. %
This seems to hold true for both ACL and INLG metrics.

\section{Discussion}
\label{sec:discussion}

Our survey reveals both positive and negative aspects of current trends in NLG evaluation. Undoubtedly positive is the fact that the vast majority of researchers do make their code and resources available after publication, despite no obligation to do so. Additionally, it is encouraging to see that types of metrics used differ given the task, suggesting an effort to use metrics which are relevant to the research goals. Overlap metrics are mostly complemented by metrics from other categories (cf.\ Figure~\ref{fig:coocurrence_categories} in the Appendix).

On the other hand, the predominance of Overlap metric types is concerning given their well-known caveats, such as their inability to measure faithfulness and poor correlation with human judgements \citep{reiter-2018-structured}. This is also compounded by the tendency to not state the rationale for the use of a metric. Without any rationale of why a given metric or set of metrics are being used, there is uncertainty on what researchers are looking to measure and whether they chose the right metrics. Our survey also reveals an over-reliance on reference-based metrics. This might be a hold-over from when generation tasks were more highly constrained and focused on more closed-domain problems such as weather forecast generation, with a defined set of reference ``gold-standard'' corpora. However, most generation problems are increasingly open-ended and require accepting a wider range of outputs that are not possible to cover in a given reference set. Therefore, it is possible that an attitudinal or structural change is needed within the research community to ask deeper questions on the use of inappropriate metrics.     

Another observation is the relationship between automatic and human evaluations. Out of the metrics that had no rationale provided, around half performed human evaluations, yet did not investigate any link between the automatic and human evaluation results. This suggest that the majority of researchers treat their evaluations as separate entities. However, given the overall lack of rationales provided for the use of automatic metrics, we cannot be certain that authors are looking to measure different aspects with their automatic and human evaluations or whether the evaluations are in fact intended to be complementary. Ultimately, this creates uncertainty for researchers reading papers and makes the reproduction of evaluations challenging.

\section{Recommendations}
\label{sec:recommendations}

\subsection{Evaluation Quality}

\paragraph{Rationalize your selection of metrics}
Authors should consider the appropriateness of the metrics they are using and whether adding more automatic metrics will in fact yield interesting insights.
In particular, we advise authors to state clearly what they expect to evaluate with each given metric so that 
there is clarity for those trying to interpret reported results.
In our investigation, we found that less than 13\% of metric occurrences are supported by a rationale other than following previous work.
Rationales are also important due to the number of metrics used -- 283 unique metrics were used at the surveyed venues last year.
We cannot reasonably expect readers to be familiar with all of them, which strengthens the need for justification.

\paragraph{Do not copy-paste widely used metrics}
We found that around 10\% of metric usages (and an unknown portion of the 77\% with no rationale provided) are justified on the grounds that they follow evaluations done in previous work.
Authors should question whether these metrics truly measure the intended qualities in the evaluation, and if they do, the authors should share their reasoning in the paper.
However, if the metrics fail to show a correlation with human judgment or a specific quality, we strongly advise authors to omit them, or at least relegate them into the appendix to clearly show their decreased priority.

\paragraph{Comment on metric combinations}
Given that automatic metrics frequently have blind spots, we also recommend commenting on the chosen combination of metrics: how do the metrics complement each other to provide a more objective evaluation of a system?

\paragraph{Respect the intended use of metrics}
Generally, when a new metric is proposed, its authors demonstrate its suitability for a given setting or task.
However, we frequently see metrics used for purposes that they were not intended for.
In such a case, the authors should justify their use of the metric from first principles or empirically.

\paragraph{Discuss (dis)agreements between human and automatic evaluation}
For both automatic and human evaluations, it is important to state the similarities or differences between their measurements. Where there are overlaps in what is being measured, authors should consider commenting on whether they see correlations between the reported results or not.

\subsection{Reproducibility}
\paragraph{Share evaluation details}
When using a library implementation of an automatic metric, the authors should first and foremost disclose which library was used -- this happened for only 34.2\% of the metrics used at INLG and 42.6\% at ACL.
Furthermore, it is also desirable to share in the appendix the parameters used to obtain the results.
Such parameters can include the version of the library, the tokenizer, the preprocessing methods, and so on.
Even better, some libraries, such as SacreBLEU \citep{post-2018-call} include easily shareable version strings with the encoding of these parameters.

\paragraph{Share data samples}
The lack of error analyses conducted within the NLG research community is a known problem \cite{van-miltenburg-etal-2023-barriers}, given the lack of comprehensiveness of both automatic and human evaluations. If possible, authors should consider sharing example outputs with metric results and adding human annotations (if a human annotation has been performed).

Additionally, we encourage the authors to release the full datasets with the evaluated system outputs. As a result, the future authors will have the possibility of using other, possibly new metrics to compare to their new systems.

\paragraph{Release code}
The final set of recommendations relate to provision of experimental code and resources. 
While code is often provided now, practices still vary considerably.
Improvements include not just releasing the code for the evaluations conducted, but also giving appropriate installation instructions and describing how the code relates to results in the paper. The inclusion of generated outputs enables evaluation reproductions and allows future evaluations with newer or alternative metrics. Finally, a structural improvement that the research community could consider is to make code and resources a requirement, subject to validation, with the camera-ready version of an accepted paper.

\section{Conclusion}
\label{sec:conclusions}

We have presented our analyses and a new dataset of 102 papers annotated with nine attributes to ascertain the different metrics, used currently by authors in NLG across publications in 2023 in both INLG and ACL venues. The process of creating and validating the annotation schema, the analyses that we have conducted, and the results we have obtained are described in this paper. 

From the results that we have obtained, we have shown that there are outstanding issues related to the type and number of metrics used, the lack of comparison and linkage between automatic and human evaluation results, and missing justifications for the selection of metrics. 

We have proposed several recommendations in the hope to offer possible solutions to these structural problems. However, while many papers have or will make recommendations on improving evaluation practices, it is only when these solutions are adopted that we as a research field can make progress on these issues.  

Our main conclusion is, that as a field, we need to provide more information on the usage of automatic metrics and the motivations behind their usage. Only by doing this can we start to bring more clarity to how evaluations are being conducted and help to alleviate adjacent challenges such as the reproduction and repeatability of evaluations.

\section*{Limitations}
\label{sec:limitations}
While this work provides a snapshot of automatic evaluation practices in NLG during 2023, quantitatively capturing long-term trends in these practices was out of the scope of this work.

\section*{Ethics Statement}
\label{sec:ethics}

The focus of this work is to gain better insights into automatic evaluation practices. The annotations made in this paper were made by the authors and therefore we did not recruit any external annotators nor process any personal data.

\section*{Supplementary Materials Availability Statement} The code for the analysis, the selection of papers, and the annotations presented in this paper are made available from our GitHub project repository at \url{https://github.com/patuchen/nlg_metric_usage}.

\section*{Acknowledgements}

This work was co-funded by the European Union (ERC, NG-NLG, 101039303) and Charles University projects GAUK 40222 and SVV 260~698.

\bibliography{nlg-metrics, anthology}

\clearpage
\appendix

\section*{Appendices}
\label{sec:appendix}
\section{Additional Results}\label{app:moreresults}
\subsection{BLEU and ROUGE Variants}
\begin{figure}[t]
    \centering
    \includegraphics[width=\linewidth]{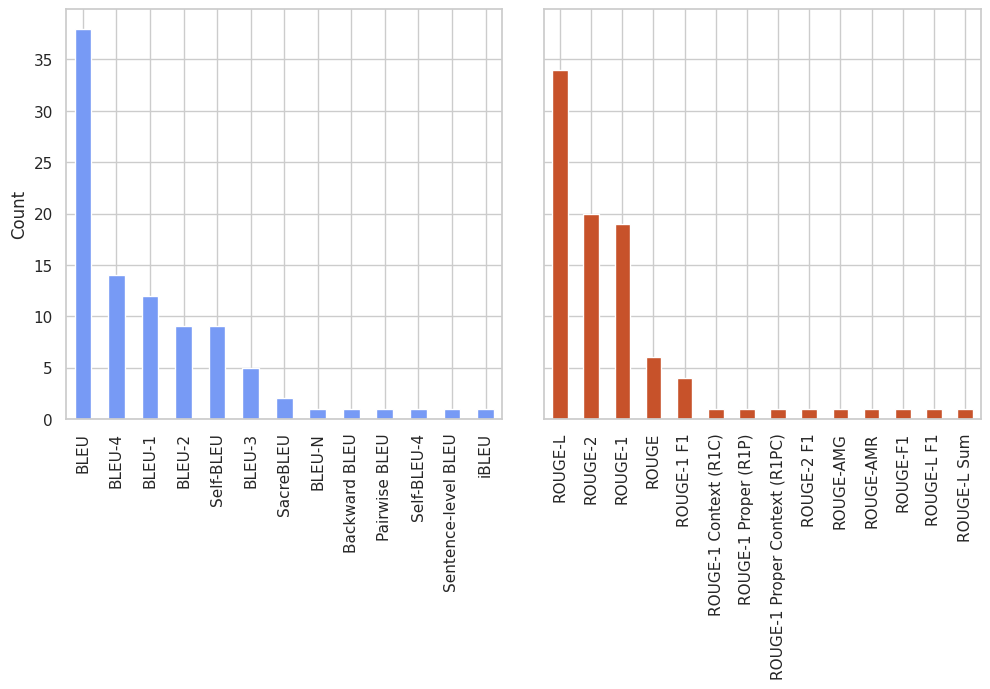}
    \caption{BLEU and ROUGE variant counts across INLG and ACL papers}
    \label{fig:BLEU_and_ROUGE}
\end{figure}

Figure \ref{fig:BLEU_and_ROUGE} shows the distribution of the different variants of BLEU and ROUGE respectively used by researchers across both INLG and ACL papers. 

\subsection{Evaluation Rationales}

\begin{figure}
    \centering
    \includegraphics[width=\linewidth]{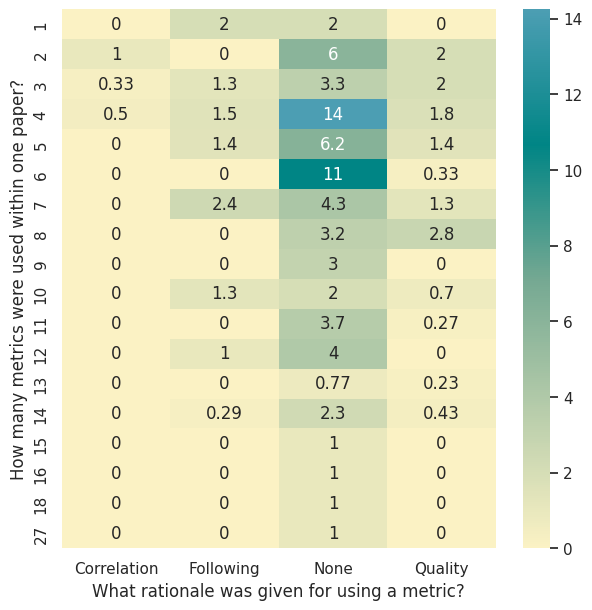}
    \caption{Number of metrics per paper against the rationale type given. If a paper provided more than one type of rationale, its contribution was proportionally divided into more categories.}
    \label{fig:metric_paper_rationale}
\end{figure}

Figure \ref{fig:metric_paper_rationale} provides a granular view of the number of metrics per paper against the rationale type given.  
We can see that correlation with human judgment is only used as a rationale when there are less metrics (2-4). 
Furthermore, if authors use 9 or more metrics, they rarely provide some insight into why the metrics were chosen.

\subsection{Metric Category Co-occurrences}
\begin{figure}[t]
    \centering
    \includegraphics[width=\linewidth]{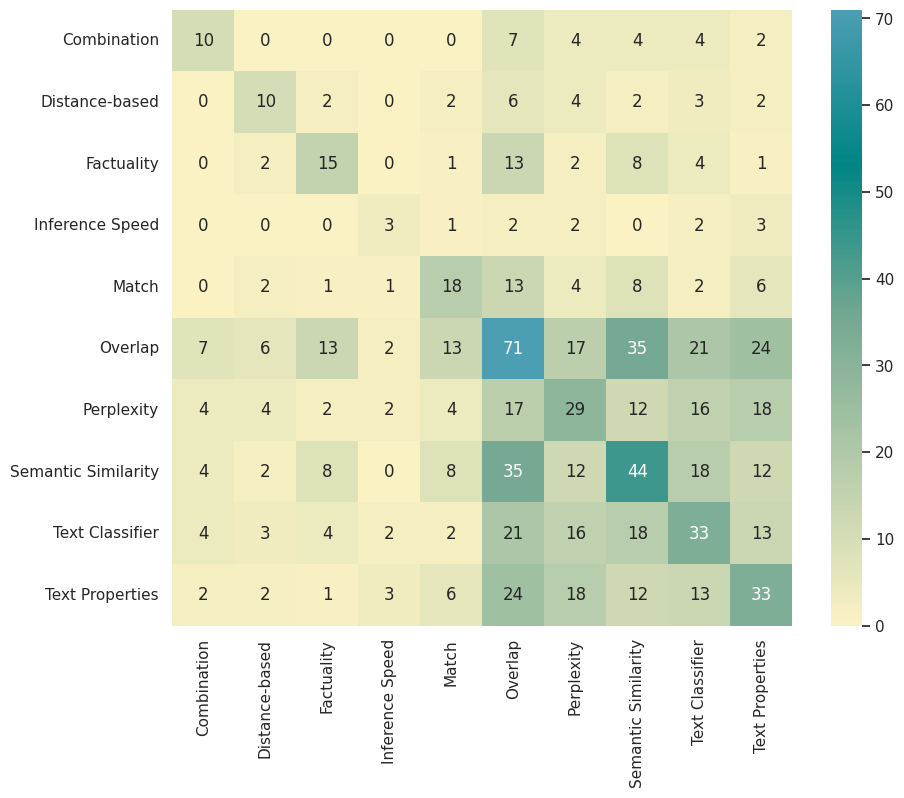}
    \caption{Co-occurrence of metric categories within papers.}
    \label{fig:coocurrence_categories}
\end{figure}

Figure \ref{fig:coocurrence_categories} supports the finding that Overlap metrics are generally used with another type of metric.

\section{List of NLG Tasks}\label{app:tasks}
The following is the list of NLG (sub-)tasks commonly mentioned in the annotated papers. Annotators were also able to note tasks not in this list.
\begin{compactitem}
    \item aggregation
    \item compression / lossy simplification
    \item content ordering/structuring
 \item content selection/determination
\item data-to-text generation
\item deep generation (DLR to text)
\item dialogue turn generation
\item end-to-end text generation
\item feature-controlled generation
\item lexicalisation
\item machine translation
\item paraphrasing / lossless simplification
\item question answering
\item question generation
\item referring expression generation
\item summarisation (text-to-text)
\item surface realisation (SLR to text)
\end{compactitem}

The following tasks were added during the annotation:
\begin{compactitem}
    \item story generation  
    \item language model sampling
    \item song lyric generation
    \item commonsense reasoning
\end{compactitem}

\section{Evaluation Metrics Used in the Annotated Papers}\label{app:metrics}
In this section, we present all of the metrics we encountered during our annotation process.
We assigned a family (fine-grained) and a category (high-level) to each metric to increase the clarity of presented results.
In some cases, e.g. for `Combination', family and type are identical.
Similarly, if a metric is prevalent, it can be in its own singleton family.

\subsection{Combination}
Multiple metrics in a simple (e.g. mean) or trained combination.
\begin{compactitem}
    \item AUC \citep{AUC_BRADLEY19971145}
    \item Average \citep{gu-etal-2023-controllable}
    \item Average of ROUGE-1, ROUGE-2, and~ROUGE-L \citep{calderon-etal-2023-systematic}
    \item BLEU area under curve \citep{meister-etal-2023-natural}
    \item G-score \citep{han-etal-2023-text}
    \item GeomMean(.) \citep{yang-jin-2023-attractive}
    \item GeomMean(Acc,Sim,Fl) \citep{jia-etal-2023-sample}
    \item Harmonic Mean of Pairwise BLEU and BLEU \citep{e-etal-2023-divhsk}
    \item HAUSER Quality \citep{he-etal-2023-hauser}
    \item J(Acc,Sim,Fl) \citep{jia-etal-2023-sample}
\end{compactitem}

\subsection{Distance-based}
Metrics that measure the distance between two distributions or sequences.

\subsubsection{Distribution Comparison}
Metrics that measure the distance between two distributions.
\begin{compactitem}
    \item Forward KL divergence of learned distribution \citep{meister-etal-2023-efficacy}
    \item Jensen-Shannon divergence of learned distribution \citep{meister-etal-2023-efficacy}
    \item Reverse cross-entropy of learned distribution \citep{meister-etal-2023-efficacy}
    \item Reverse KL divergence of learned distribution \citep{meister-etal-2023-efficacy}
    \item Total variation distance of learned distribution \citep{meister-etal-2023-efficacy}
    \item Weighted macro-F1 \citep{meister-etal-2023-efficacy}
    \item Zipf's Coefficient \citep{han-etal-2023-ssd}
\end{compactitem}

\subsubsection{Edit Distance}
Metrics that measure the edit distance between two sequences.
\begin{compactitem}
    \item $D_{lex}$ \citep{li-etal-2023-explicit}
    \item $D_{syn}$ \citep{li-etal-2023-explicit}
    \item Edit Distance \citep{ippolito-etal-2023-preventing}
    \item $Pres_{COMB}$ \citep{gao-etal-2023-rarr}
    \item TER \citep{li-etal-2023-shot-data, zandie-etal-2023-cogen}
\end{compactitem}

\subsubsection{Loss/Error}
Metrics that measure the loss or error between the generated output and a gold reference.

\begin{compactitem}
    \item Agreement - the number of questions generated by GPT-2 (\#Q) matches the number of GPT-3 annotated questions for a given problem \citep{shridhar-etal-2023-distilling}
    \item Bias \citep{pei-etal-2023-preadd}
    \item Cropped sentence ratio \citep{tian-etal-2023-unsupervised}
\end{compactitem}

\subsection{Factuality (Category)}
Metrics that either directly or indirectly aim to measure factuality.

\subsubsection{Factuality (Family)}
Metrics that either directly aim to measure factuality.
\begin{compactitem}
    \item AlignScore \citep{zha-etal-2023-alignscore}
   \item CheXpert factuality \citep{trienes-etal-2023-guidance}
   \item Content Selection \citep{thomson-etal-2023-enhancing}
   \item DecompEval \citep{ke-etal-2023-decompeval}
   \item FactCC \citep{kryscinski-etal-2020-evaluating}
   \item FEQA \citep{zha-etal-2023-alignscore}
   \item NEHR \citep{akani-etal-2023-reducing}
   \item NER Overlap \citep{zha-etal-2023-alignscore}
   \item $Q^2$ \citep{wu-etal-2023-wecheck}
   \item QAFactEval \citep{zha-etal-2023-alignscore,wu-etal-2023-wecheck}
   \item QuestEval \citep{zha-etal-2023-alignscore,wu-etal-2023-wecheck}
   \item Relation Generation \citep{thomson-etal-2023-enhancing}
   \item WeCheck \citep{wu-etal-2023-wecheck}
\end{compactitem}

\subsubsection{NLI}
Classifiers into three classes: logical entailment, contradiction, and neutrality.

\begin{compactitem}
    \item ANLI \citep{wu-etal-2023-wecheck,narayan-etal-2023-conditional}
    \item $Attr_{AUTO}$ \citep{gao-etal-2023-rarr}
    \item DeBERTaxxlargev2 \citep{hirsch-etal-2023-revisiting}
    \item NLI \citep{garneau-lamontagne-2023-guided,li-etal-2023-shot-data}
    \item NLI-warmup  \citep{wu-etal-2023-wecheck}
    \item NUBIA Agreement \citep{kane-etal-2020-nubia}
    \item NUBIA Contradiction \citep{kane-etal-2020-nubia}
    \item NUBIA Neutrality \citep{kane-etal-2020-nubia}
    \item P-NLI \citep{zeng-etal-2023-seen}
    \item SUMMAC \citep{wu-etal-2023-wecheck}
\end{compactitem}

\subsection{Inference Speed}
Metrics that measure the inference speed of a model.
\begin{compactitem}
    \item Inference Time \citep{kumar-etal-2023-controlled}
    \item Latency \citep{huang-etal-2023-directed}
    \item Speed (token per s) \citep{liu-etal-2023-bolt}
    \item Throughput \citep{huang-etal-2023-directed}
\end{compactitem}

\subsection{Match}
Metrics that measure the match between a generated output and a gold label.

\subsubsection{Accuracy}
Metrics that measure accuracy.
\begin{compactitem}
    \item Accuracy
    \item Accuracy of comparator \citep{yang-etal-2023-fantastic}
    \item Accuracy of keyword inclusion \citep{sasazawa-etal-2023-controlling}
    \item Accuracy of keyword inclusion at a specified position \citep{sasazawa-etal-2023-controlling}
    \item Accuracy of vehicle \citep{yang-etal-2023-fantastic}
    \item Completion Sensitivity Score \citep{sieker-etal-2023-beyond}
    \item Domain Accuracy \citep{liu-etal-2023-system-initiated}
    \item Domain Slot Value Accuracy \citep{liu-etal-2023-system-initiated}
    \item Exact Match \citep{tang-etal-2023-mvp}
    \item Exact Match Accuracy \citep{skitalinskaya-etal-2023-claim}
    \item Inform \citep{tang-etal-2023-mvp}
    \item Proportion of sentences with comparator words \citep{yang-etal-2023-fantastic}
    \item Stress-duration alignment \citep{tian-etal-2023-unsupervised}
    \item Success \citep{tang-etal-2023-mvp}
    \item Transition Accuracy \citep{liu-etal-2023-system-initiated}
\end{compactitem}

\subsubsection{F1}
Metrics that measure F1.
\begin{compactitem}
    \item F1
    \item F1 (Lexical Simplification) \citep{sun-etal-2023-teaching}
    \item F1-score (appraisal) \citep{menchaca-resendiz-klinger-2023-affective}
    \item Format F1 \citep{qian-etal-2023-unilg}
    \item Knowledge-F1 \citep{huang-etal-2023-directed}
    \item macro-F1 \citep{same-etal-2023-models,feng-etal-2023-dunst}
    \item micro-F1 \citep{xu-etal-2023-unsupervised}
    \item QA-F1 (informativeness/grounding) \citep{narayan-etal-2023-conditional}
    \item weighted macro-F1 \citep{same-etal-2023-models}
\end{compactitem}

\subsubsection{Precision}
Metrics that measure precision.
\begin{compactitem}
    \item Knowledge-Precision \citep{huang-etal-2023-directed}
    \item Precision \citep{same-etal-2023-models}
    \item Precision (Lexical Simplification) \citep{sun-etal-2023-teaching}
\end{compactitem}

\subsubsection{Recall}
Metrics that measure recall.
\begin{compactitem}
    \item Knowledge-Recall \citep{huang-etal-2023-directed}
    \item Local Recall \citep{van-der-lee-etal-2023-neural}
    \item Recall \citep{same-etal-2023-models,li-etal-2023-language-modeling}
    \item Recall (Lexical Simplification) \citep{sun-etal-2023-teaching}
    \item Recall@N \citep{hwang-etal-2023-gan}
\end{compactitem}

\subsection{Overlap (Category)}
Metrics that measure the overlap between two sequences.

\subsubsection{BLEU}
Multiple variants of the BLEU score \citep{papineni-etal-2002-bleu}.
\begin{compactitem}
    \item Backward BLEU \citep{xie-etal-2023-next}
    \item BLEU \citep{papineni-etal-2002-bleu}
    \item BLEU-1 \citep{papineni-etal-2002-bleu}
    \item BLEU-2 \citep{papineni-etal-2002-bleu}
    \item BLEU-3 \citep{papineni-etal-2002-bleu}
    \item BLEU-4 \citep{papineni-etal-2002-bleu}
    \item BLEU-N \citep{wu-etal-2023-focus}
    \item iBLEU \citep{li-etal-2023-explicit}
    \item Pairwise BLEU \citep{e-etal-2023-divhsk}
    \item SacreBLEU \citep{post-2018-call}
    \item Self-BLEU (between source and target) \citep{zhu2018texygen}
    \item Self-BLEU (between more system-generated outputs) \citep{zhu2018texygen}
    \item Self-BLEU-4 \citep{he-etal-2023-diffusionbert}
    \item Sentence-level BLEU \citep{tian-etal-2023-unsupervised}
\end{compactitem}

\subsubsection{chrF++}
This family consists solely of the chrF++ metric \citep{popovic-2015-chrf}.

\subsubsection{CIDEr}
This family consists solely of the CIDEr metric \citep{Vedantam_CIDEr_2015_CVPR}.

\subsubsection{METEOR}
This family consists solely of the METEOR metric \citep{banerjee-lavie-2005-meteor}.

\subsubsection{NIST}
Multiple variants of the NIST metric \citep{Doddington_NIST}.

\begin{compactitem}
    \item NIST \citep{Doddington_NIST}
    \item NIST-1 \citep{tang-etal-2023-enhancing}
    \item NIST-2 \citep{tang-etal-2023-enhancing}
    \item NIST-3 \citep{tang-etal-2023-enhancing}
    \item NIST-4 \citep{tang-etal-2023-enhancing}
\end{compactitem}

\subsubsection{Overlap (family)}
Metrics that measure the overlap between two sequences.
\begin{compactitem}
	\item Add \citep{sun-etal-2023-teaching}
	\item Copy Success Rate (word) \citep{huang-etal-2023-extensible}
	\item Coverage \citep{van-der-lee-etal-2023-neural,li-etal-2023-cats}
	\item Coverage (of keywords) \citep{liu-etal-2023-bolt}
	\item D-add \citep{sun-etal-2023-teaching}
	\item D-delete \citep{sun-etal-2023-teaching}
	\item Delete \citep{sun-etal-2023-teaching}
	\item Dkeep \citep{sun-etal-2023-teaching}
	\item Extractive fragment density ($\rho$) \citep{mascarell-etal-2023-entropy}
	\item HAUSER Creativity \citep{he-etal-2023-hauser}
	\item Keep \citep{sun-etal-2023-teaching}
	\item MS-Jaccard \citep{xie-etal-2023-next}
	\item Phonetic Overlap \citep{loakman-etal-2023-twistlist}
	\item Proper Noun Ratio (P Ratio) \citep{chang-etal-2023-revisiting-architectures}
	\item Salient word coverage \citep{tian-etal-2023-unsupervised}
	\item Slot Coverage \citep{surya-etal-2023-zero}
	\item SMART \citep{cripwell-etal-2023-context}
	\item Weisfeiler Lehman graph hash \citep{bhandari-brennan-2023-trustworthiness}
\end{compactitem}

\subsubsection{PARENT}
Multiple scores produced by the PARENT metric \citep{dhingra-etal-2019-handling}.
\begin{compactitem}
    \item PARENT\citep{dhingra-etal-2019-handling}
    \item PARENT-T-F1 \citep{huang-etal-2023-directed}
    \item PARENT-T-Precision \citep{huang-etal-2023-directed}
    \item PARENT-T-Recall \citep{huang-etal-2023-directed}
\end{compactitem}

\subsubsection{ROUGE}
Multiple variants of the ROUGE score \citep{lin-2004-rouge}.

\begin{compactitem}
	\item ROUGE \citep{lin-2004-rouge}
	\item ROUGE-1 \citep{lin-2004-rouge}
	\item ROUGE-1 Context (R1C) \citep{chang-etal-2023-revisiting-architectures}
	\item ROUGE-1 F1 \citep{chang-etal-2023-revisiting-architectures}
	\item ROUGE-1 Proper (R1P) \citep{chang-etal-2023-revisiting-architectures}
	\item ROUGE-1 Proper Context (R1PC) \citep{chang-etal-2023-revisiting-architectures}
	\item ROUGE-2 \citep{lin-2004-rouge}
	\item ROUGE-2 F1 \citep{jia-etal-2023-reducing}
	\item ROUGE-AMG  \citep{juan-etal-2023-generating}
	\item ROUGE-AMR \citep{juan-etal-2023-generating}
	\item ROUGE-F1 \citep{huang-etal-2023-summaries}
	\item ROUGE-L \citep{lin-2004-rouge}
	\item ROUGE-L F1 \citep{jia-etal-2023-reducing}
	\item ROUGE-L Sum \citep{narayan-etal-2023-conditional}
\end{compactitem}

\subsubsection{SARI}
Two scores produced by the SARI metric \citep{xu-etal-2016-optimizing}.
\begin{compactitem}
    \item DSARI \citep{xu-etal-2016-optimizing}
    \item SARI \citep{xu-etal-2016-optimizing}
\end{compactitem}

\subsection{Perplexity (Category)}
Metrics that directly or indirectly measure perplexity.

\subsubsection{MAUVE}
MAUVE metric \citep{pillutla-etal:mauve:neurips2021} with various underlying language models.

\begin{compactitem}
    \item MAUVE \citep{pillutla-etal:mauve:neurips2021}
    \item MAUVE (ELECTRA-large) \citep{he-etal-2023-blind}
    \item MAUVE (GPT2-large) \citep{he-etal-2023-blind}
    \item MAUVE (RoBERTa-large) \citep{he-etal-2023-blind}
\end{compactitem}

\subsubsection{Perplexity (family)}
Metrics that directly measure perplexity.
\begin{compactitem}
	\item Bits per character (BPC) \citep{nawrot-etal-2023-efficient}
	\item Fluency \citep{pei-etal-2023-preadd}
	\item Fluency (Perplexity) \citep{yang-jin-2023-attractive}
	\item GPT-PPL \citep{he-etal-2023-blind}
	\item MLM-PPL \citep{he-etal-2023-blind}
	\item Model PPL \citep{feng-etal-2023-dunst}
	\item Output PPL \citep{feng-etal-2023-dunst}
	\item Perplexity
	\item Perplexity \citep{liang-etal-2023-open,tang-etal-2023-mvp}
	\item Perplexity (Chinese GPT-2) \citep{yang-etal-2023-fantastic}
	\item Perplexity (GPT-2)  \citep{tian-etal-2023-unsupervised}
\end{compactitem}

\subsection{Semantic Similarity (Category)}
Metrics that measure semantic similarity.

\subsubsection{BARTScore}
Multiple scores produced by the BARTScore metric \citep{yuan-2021-bartscore}.
\begin{compactitem}
	\item BARTScore\citep{yuan-2021-bartscore}
	\item BARTScore faithfulness \citep{he-etal-2023-blind}
	\item BARTScore fscore \citep{he-etal-2023-blind}
	\item BARTScore precision \citep{he-etal-2023-blind}
	\item BARTScore recall \citep{he-etal-2023-blind}
\end{compactitem}

\subsubsection{BERTScore}
Multiple scores produced by the BERTScore metric \citep{zhang-2019-bertscore}.
\begin{compactitem}
	\item BERTScore \citep{zhang-2019-bertscore}
	\item BERTScore F1 \citep{zhang-2019-bertscore}
	\item BERTScore Precision \citep{zhang-2019-bertscore}
	\item BERTScore Recall \citep{zhang-2019-bertscore}
\end{compactitem}

\subsubsection{MoverScore}
This family consists solely of the MoverScore metric \citep{zhao-etal-2019-moverscore}.

\subsubsection{Semantic Similarity (family)}
Metrics that directly measure semantic similarity.

\begin{compactitem}
	\item Coherence \citep{li-etal-2023-language-modeling, li-etal-2023-contrastive}
	\item Cosine Similarity \citep{chung-etal-2023-increasing}
	\item Embedding Similarity \citep{mukherjee-dusek-2023-leveraging}
	\item MPNet Cosine Similarity \citep{anschutz-etal-2023-correct}
	\item NegMPNet Cosine Similarity \citep{anschutz-etal-2023-correct}
	\item NUBIA Semantic Similarity \citep{kane-etal-2020-nubia}
	\item P-SIM \citep{zeng-etal-2023-seen}
	\item RANK \citep{garneau-lamontagne-2023-guided}
	\item Relevance \citep{pei-etal-2023-preadd}
	\item Semantic Similarity \citep{jia-etal-2023-sample}
	\item Sentence-BERT \citep{surya-etal-2023-zero}
	\item Sentence-BERT Cosine Similarity \citep{jing-etal-2023-multi}
	\item SimCSE \citep{zha-etal-2023-alignscore}
	\item Spearman Rank Correlation \citep{hwang-etal-2023-gan}
	\item SR (Semantic Repetition) \citep{liang-etal-2023-open}
	\item Topic modelling \citep{bhandari-brennan-2023-trustworthiness}
\end{compactitem}

\subsection{Text Classifiers}
Type of metrics that classify various properties of the generated text.

\subsubsection{BLEURT}
Metrics based on BLEURT \citep{sellam-2020-bleurt}.

\begin{compactitem}
    \item BLEURT \citep{sellam-2020-bleurt}
    \item NegBLEURT \citep{anschutz-etal-2023-correct}
    \item Purity Score \citep{cafagna-etal-2023-hl}
\end{compactitem}

\subsubsection{Quality Estimation}
Quality estimation metrics for referenceless evaluation. Also includes a small set of classifiers trained to distinguish human-written from machine-generated texts.
\begin{compactitem}
    \item BERT Classification F1 \citep{almasi-schionning-2023-fine}
	\item BERT Classification Precision \citep{almasi-schionning-2023-fine}
	\item BERT Classification Recall \citep{almasi-schionning-2023-fine}
	\item BLANC \citep{zha-etal-2023-alignscore}
	\item COMET-QE \citep{he-etal-2023-blind}
	\item CTC \citep{nimah-etal-2023-nlg}
	\item CTRLEval \citep{ke-etal-2022-ctrleval}
	\item GPTRank \citep{jiang-etal-2023-llm}
    \item LR Classification F1 \citep{almasi-schionning-2023-fine}
	\item LR Classification Precision \citep{almasi-schionning-2023-fine}
	\item LR Classification Recall \citep{almasi-schionning-2023-fine}
	\item Naturalness \citep{narasimhan-etal-2023-text}
	\item PRISM-QE \citep{he-etal-2023-blind}
	\item USR \citep{ke-etal-2023-decompeval}
\end{compactitem}

\subsubsection{Style Classifiers}
Classifiers that were trained to classify style, sentiment, or topic.
\begin{compactitem}
	\item Accuracy (Sentiment) \citep{huang-etal-2023-extensible}
	\item Accuracy (Tense) \citep{huang-etal-2023-extensible}
	\item Accuracy (Topic) \citep{huang-etal-2023-extensible}
	\item Act - Classification accuracy (A-ACC) Roberta \citep{zeng-etal-2023-seen}
	\item Act - Multiple Attribute Evaluation (A-MAE) \citep{zeng-etal-2023-seen}
	\item Bias (absolute value of relevance - 50) \citep{ma-etal-2023-focused}
	\item C-Ext \citep{han-etal-2023-ssd}
	\item Content Ordering \citep{thomson-etal-2023-enhancing}
	\item Correctness \citep{yang-etal-2023-tailor}
	\item custom trained relevance classifier \citep{ma-etal-2023-focused}
	\item Detoxify \citep{bhandari-brennan-2023-trustworthiness}
	\item Emotion - Classification accuracy (E-ACC) Roberta \citep{zeng-etal-2023-seen}
	\item Emotion - Multiple Attribute Evaluation (E-MAE) \citep{zeng-etal-2023-seen}
	\item Fluency \citep{jia-etal-2023-sample}
	\item Grammaticality \citep{kim-etal-2023-critic,xu-etal-2023-best,yang-etal-2023-tailor}
	\item Integrity \citep{qian-etal-2023-unilg}
	\item Intented Sentiment (external classifier) \citep{liu-etal-2023-bolt}
	\item Intented Sentiment (internal classifier)\citep{liu-etal-2023-bolt}
	\item Label Accuracy \citep{chung-etal-2023-increasing}
	\item LENS \citep{maddela-etal-2023-lens}
	\item Negative Sentiment \citep{kumar-etal-2023-controlled}
	\item P-Multiple Attribute Evaluation (P-MAE) \citep{zeng-etal-2023-seen}
	\item Positiveness \citep{kim-etal-2023-critic}
	\item RoBERTa fine-tuned for sentiment \citep{ma-etal-2023-focused}
	\item Sentiment \citep{gu-etal-2023-controllable}
	\item Sentiment Accuracy \citep{han-etal-2023-text,mukherjee-dusek-2023-leveraging}
	\item Simile confidence \citep{yang-etal-2023-fantastic}
	\item Simplicity \citep{kumar-etal-2023-controlled}
	\item Structure F1 \citep{qian-etal-2023-unilg}
	\item Style Accuracy \citep{yang-jin-2023-attractive, jia-etal-2023-sample}
	\item Style Transfer Accuracy \citep{narasimhan-etal-2023-text}
	\item Success  \citep{pei-etal-2023-preadd}
	\item Topic \citep{gu-etal-2023-controllable}
	\item Toxicity  \citep{pei-etal-2023-preadd, kim-etal-2023-critic}
	\item Toxicity \citep{kumar-etal-2023-controlled}
	\item $\Delta$ TextBlob \citep{sheng-etal-2023-learning}
\end{compactitem}

\subsubsection{Unieval}
Various scores produced by the Unieval metric \citep{zhong-etal-2022-towards}.
\begin{compactitem}
	\item UniEval \citep{zhong-etal-2022-towards}
	\item Unieval - coherence  \citep{he-etal-2023-blind}
	\item Unieval - consistency \citep{he-etal-2023-blind}
	\item Unieval - fluency \citep{he-etal-2023-blind}
	\item Unieval - overall \citep{he-etal-2023-blind}
	\item Unieval - relevance \citep{he-etal-2023-blind}
	\item UniEval (Dial) \citep{ke-etal-2023-decompeval}
	\item UniEval (Summ) \citep{ke-etal-2023-decompeval}
\end{compactitem}

\subsection{Text Properties}
Type of metrics that measure various text properties.

\subsubsection{Flesch Readability}
Flesch Readability scores.
\begin{compactitem}
	\item Flesch Reading Ease Score \citep{bhandari-brennan-2023-trustworthiness}
    \item Flesch-Kincaid grade level (FKGL) \citep{Flesch1948}
\end{compactitem}

\subsubsection{N-gram Diversity}
N-gram diversity metrics.
\begin{compactitem}
	\item Averaged Distinctiveness \citep{huang-etal-2023-extensible}
	\item Bigram TTR \citep{van-der-lee-etal-2023-neural}
	\item Dist-n \citep{feng-etal-2023-dunst}
	\item Distinct-1 \citep{li-etal-2016-diversity}
	\item Distinct-2 \citep{li-etal-2016-diversity}
	\item Distinct-3 \cite{see-etal-2019-massively}
	\item Distinct-3 (proportion) \citep{liu-etal-2023-bolt}
	\item Distinct-4  \citep{tang-etal-2023-mvp}
	\item Distinct-n \citep{surya-etal-2023-zero}
	\item Distinctness \citep{gu-etal-2023-controllable}
	\item Div-4 \citep{he-etal-2023-diffusionbert}
	\item Diversity \citep{li-etal-2023-language-modeling, li-etal-2023-contrastive}
	\item Diversity (of questions) \citep{juan-etal-2023-generating}
	\item Diversity score \citep{cafagna-etal-2023-hl}
	\item Diversity-1 \citep{xu-etal-2023-best}
	\item Diversity-2 \citep{xu-etal-2023-best}
	\item Diversity-3 \citep{xu-etal-2023-best}
	\item Ent-4 \citep{tang-etal-2023-enhancing}
	\item Initial Phonetic Overlap \citep{loakman-etal-2023-twistlist}
	\item Mean segmented type-token ratio \citep{van-der-lee-etal-2023-neural}
	\item n-gram novelty (n from 1-10) \citep{mccoy-etal-2023-much}
	\item Novelty \citep{van-der-lee-etal-2023-neural}
	\item Number of types \citep{van-der-lee-etal-2023-neural}
	\item Percentage of novel texts \citep{van-der-lee-etal-2023-neural}
	\item Syntactic Novelty (dependency role) \citep{mccoy-etal-2023-much}
	\item Syntactic Novelty (labeled dependency arc) \citep{mccoy-etal-2023-much}
	\item Syntactic Novelty (sentence level) \citep{mccoy-etal-2023-much}
	\item Unique Sentence Count \citep{xu-etal-2023-best}
	\item $\Delta$ CR \citep{hirsch-etal-2023-revisiting}
\end{compactitem}

\subsubsection{N-gram Repetition}
N-gram repetition metrics.
\begin{compactitem}
	\item 4-gram Repetition \citep{li-etal-2023-contrastive}
	\item Bigram Repetition \citep{li-etal-2023-contrastive}
	\item Lexical Repetition \citep{xie-etal-2023-next,liang-etal-2023-open}
	\item Repetition rate \citep{han-etal-2023-ssd}
	\item Trigram Repetition \citep{surya-etal-2023-zero,liu-etal-2023-bolt,li-etal-2023-contrastive}
\end{compactitem}

\subsubsection{Sequence Length}
Various measures of generated sequence length.
\begin{compactitem}
	\item Average Length \citep{sheng-etal-2023-learning}
	\item Average Sentence Length \citep{van-der-lee-etal-2023-neural}
	\item HAUSER Informativeness \citep{he-etal-2023-hauser}
	\item Length \citep{xie-etal-2023-next}
	\item Sentence Count \citep{xu-etal-2023-best}
	\item Sentence Length \citep{bhandari-brennan-2023-trustworthiness}
	\item Shortening Factor (SF) \citep{nawrot-etal-2023-efficient}
	\item Standard deviation of the sentence length \citep{van-der-lee-etal-2023-neural}
\end{compactitem}

\section{Paper and Code Resources}\label{app:code}

This section adds further detail to the results discussed in subsection~\ref{subsec:paper_resource_findings}.

\subsection{Code Releases Annotation Procedure}

\begin{figure}
    \centering
    \includegraphics[width=\linewidth]{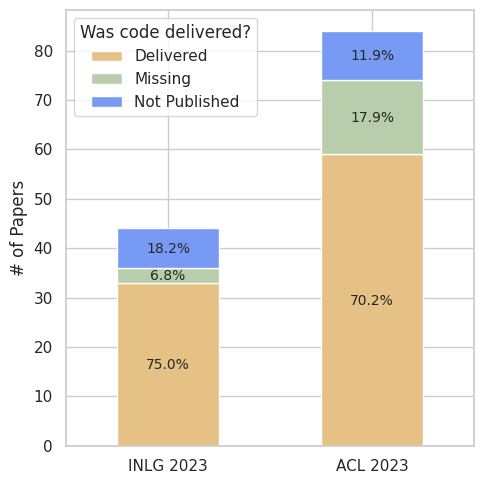}
    
    \caption{Each paper either did not link any source code (or data) or linked it and delivered or failed to deliver it -- `missing'.}
    \label{fig:delivered_mispromised} 
\end{figure}

\begin{figure}
    \centering
    \includegraphics[width=\linewidth]{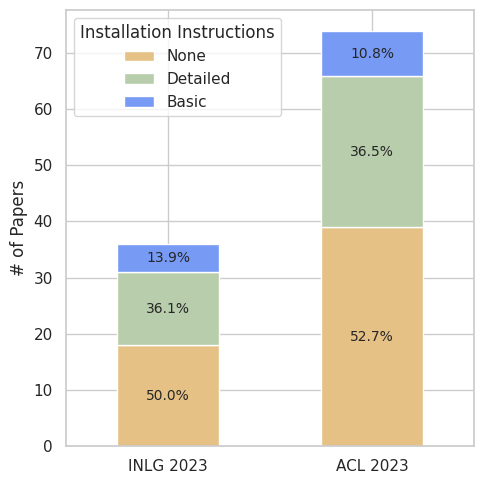}
    \caption{The quality of installation instructions annotated as \textit{None, Basic, Detailed}.}
    \label{fig:install_instr}
\end{figure}

\begin{figure}
    \centering
    \includegraphics[width=\linewidth]{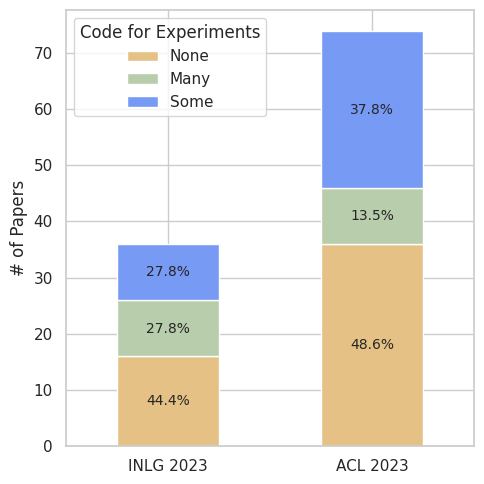}
    \caption{The quality of linking experiments in paper and code annotated as found None, Some, and Many.}
    \label{fig:experiments}
\end{figure}

\begin{figure}
    \centering
    \includegraphics[width=\linewidth]{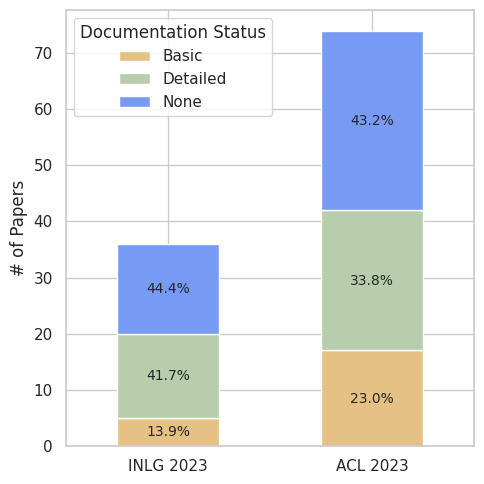}
    \caption{The quality of documentation annotated as None, Basic, Detailed.}
    \label{fig:documentation}
\end{figure}

\begin{figure}
    \centering
    \includegraphics[width=\linewidth]{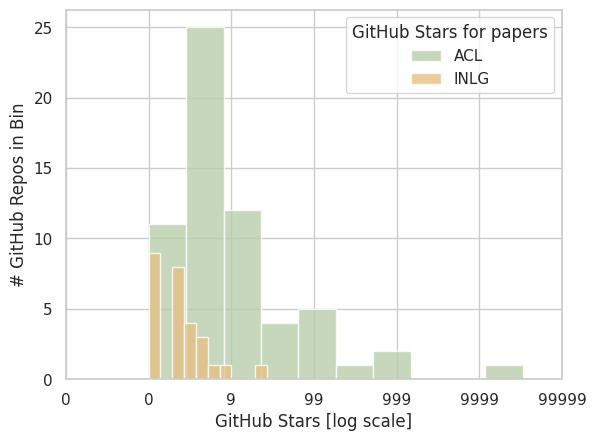}
    \caption{Distribution of GitHub Stars for INLG and ACL papers}
    \label{fig:stars_acl_inlg}
\end{figure}

\begin{figure}
    \centering
    \includegraphics[width=\linewidth]{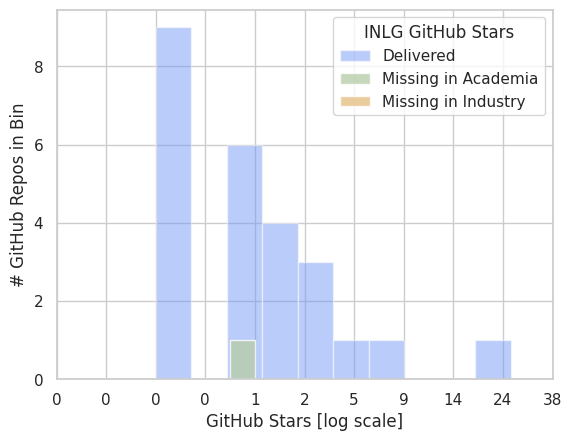}
    \includegraphics[width=\linewidth]{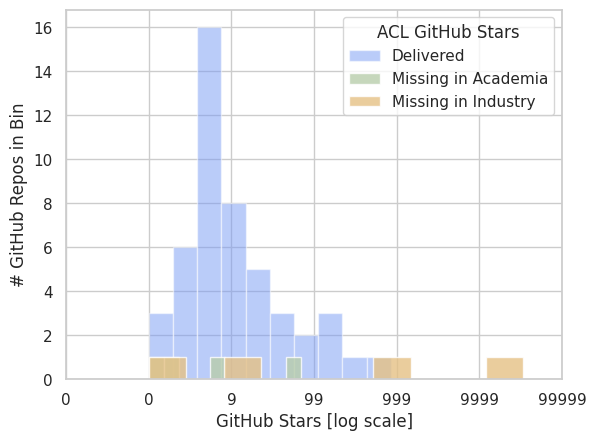}
    \caption{Distribution of the GitHub Stars for ACL for groups with groups missing for Industry and Academia}
    \label{fig:stars_mispromised}
\end{figure}

For each paper we annotated with the following procedure:
\begin{enumerate}
    \item If the paper provides a link to a code or data release.
    \item If the link actually contains the release resulting labels \textit{no code, delivered, missing)} (Figure \ref{fig:delivered_mispromised}).
    \item We annotated if the authors come from \textit{Academia} or \textit{Industry}. The mixed authoring teams received the labels \textit{Academia Industry, Industry Academia} depending on the first authors, resulting in four labels.
    \item We retrieved the GitHub Stars for each release since all except one paper was released on GitHub (Figures \ref{fig:stars_acl_inlg} and \ref{fig:stars_mispromised}).
    \item We annotated if the \textit{Installation Instructions} were provided as follows (Figure \ref{fig:install_instr}):
    \begin{itemize}
        \item None - no attempt at providing installation instructions seen.
        \item Some - installation instructions are visible but lack the necessary detail.
        \item Detailed - clearly states dependencies and exact (minimal) versions so we believe the computational environment can be easily replicated. 
    \end{itemize}
    \item We checked the clarity of the experiment structure if the experiments mentioned in the paper are \textit{ discoverable} (Figure \ref{fig:experiments}).
    \begin{itemize}
        \item \textit{None}: we have no idea how to start any experiment.
        \item \textit{Some}: we easily found how to replicate only the main experiments.
        \item \textit{Many}: we found out how to run experiments even for all the ablation groups.
    \end{itemize}
    \item We labeled the level of documentation detail with the following (Figure \ref{fig:documentation}):
    \begin{itemize}
        \item \textit{None}: no introduction to the codebase. 
        \item \textit{Basic}: it was clear what the main commands do, including the most important arguments.
        \item \textit{Detailed}: it was clear what most hyper-parameters mean and how one could change them.
    \end{itemize}
\end{enumerate}

\begin{figure}
    \centering
    \includegraphics[width=\linewidth]{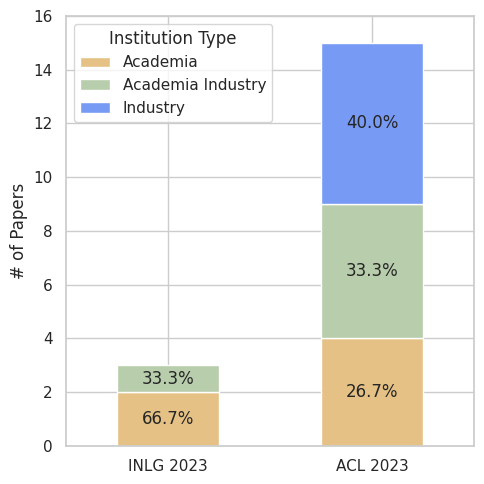}
    \caption{How the teams from academia or industry behind the papers with missing code are represented?}
    \label{fig:mispromised_academia_vs_industry}
\end{figure}

\end{document}